# Soft Computing Technique to Solve Part Family Identification Problems Using Classification and Coding System

Tamal Ghosh[*] and Pranab K Dan

Department of Industrial Engineering & Management, West Bengal University of Technology,

BF 142, Salt Lake City, Kolkata 700064, India

**Abstract.** This article deals with Part family formation problem which is believed to be moderately complicated to be solved in polynomial time in the vicinity of Group Technology (GT). In the past literature researchers investigated that the part family formation techniques are principally based on production flow analysis (PFA) which usually considers operational requirements, sequences and time. Part Coding Analysis (PCA) is merely considered in GT which is believed to be the proficient method to identify the part families. PCA classifies parts by allotting them to different families based on their resemblances in: (1) design characteristics such as shape and size, and/or (2) manufacturing characteristics (machining requirements). A novel approach based on simulated annealing namely SAPFOCS is adopted in this study to develop effective part families exploiting the PCA technique. Thereafter Taguchi's orthogonal design method is employed to solve the critical issues on the subject of parameters selection for the proposed metaheuristic algorithm. The adopted technique is therefore tested on 5 different datasets of size 5×9 to 27×9 and the obtained results are compared with C-Linkage clustering technique. The experimental results reported that the proposed metaheuristic algorithm is extremely effective in terms of the quality of the solution obtained and has outperformed C-Linkage algorithm in most instances.

# Soft Computing Technique to Solve Part Family Identification Problems Using Classification and Coding System

**Abstract.** This article deals with Part family formation problem which is believed to be moderately complicated to be solved in polynomial time in the vicinity of Group Technology (GT). In the past literature researchers investigated that the part family formation techniques are principally based on production flow analysis (PFA) which usually considers operational requirements, sequences and time. Part Coding Analysis (PCA) is merely considered in GT which is believed to be the proficient method to identify the part families. PCA classifies parts by allotting them to different families based on their resemblances in: (1) design characteristics such as shape and size, and/or (2) manufacturing characteristics (machining requirements). A novel approach based on simulated annealing namely SAPFOCS is adopted in this study to develop effective part families exploiting the PCA technique. Thereafter Taguchi's orthogonal design method is employed to solve the critical issues on the subject of parameters selection for the proposed metaheuristic algorithm. The adopted technique is therefore tested on 5 different datasets of size 5×9 to 27×9 and the obtained results are compared with C-Linkage clustering technique. The experimental results reported that the proposed metaheuristic algorithm is extremely effective in terms of the quality of the solution obtained and has outperformed C-Linkage algorithm in most instances.

## 1. Introduction

Group technology (GT) is incorporated in cellular manufacturing systems (CMS) as the manufacturing philosophy which is substantial in improving productivity for the manufacturing firms. The basic GT notion depicts that the part family identification technique is assumed to classify the parts in separate groups which require similar equipment arrangements. Group technology (GT) might be further considered as a simplified methodology which primarily groups standardized similar entities such as parts, assemblies, process plans, tools, instructions, etc. to minimize the time and cost and to improve the overall productivity for batch type production (Burbidge, 1963). As reported in the literature (Guerrero *et al.*, 2000) a successful implementation of GT could eventually minimize the engineering costs, facilitate cellular manufacturing, quicken product development, enhance costing accuracy, simplify process planning, minimize tooling cost and simplify the overall purchasing process. Therefore the major prerequisite in implementing GT is the identification of part families (Kaparthi and Suresh, 1991). A part family is a group of parts sharing similar design and manufacturing characteristics. Early research in this domain has been dedicated primarily on the formation of production-oriented part families in which similarities amongst the parts are principally recognized on the fact of their processing requirements, operation time and operation sequences. Though these methodologies are inadequate in achieving the needs of other extents of manufacturing. For example, parts with homogeneous shape, size, dimension or other design characteristics are believed to be clustered in a single family for design justification and elimination of part varieties, however parts which are clustered on the fact of homogeneous routing and the tooling needs are convenient to resolve the process planning issues.

Therefore the scope of this domain of research is believed to be expanded and examined to a wider span of part similarities. Part similarities are said to be identified sooner than the formation of part families. Part attributes such as shape, length/diameter ratio, material type, part function, dimensions, tolerances, surface finishing, process, operations, machine tool, operation sequence, annual production quantity, fixtures needed, lot sizes have been considered as the basis for similarity utilization as specified by Groover and Zimmers, (1984). Moon (1992) has stated that the complexity remains in acquiring an appropriate technique which provides an identifying competence of human being, such as identifying patterns in groups, and forming part families with the aid of intelligence.

Two different approaches are traced in past literature in order to form part families, first is production flow analysis (PFA) which deals with processing requirements of parts, operational sequences and operational time of the parts on the machines (Burbidge, 1996). Second approach is the Part Coding Analysis (PCA) which utilizes predefined coding schemes to facilitate the process using several attributes of parts such as geometrical shapes, materials, design features and functional requirements etc. (Mitrofanov, 1966).

PCA is exposed in this study as an essential and effective tool for successful implementation of GT concept. The part coding could be numbers (numerical) or alphabets (alphabetical) or a hybrid form of numbers and alphabets (alphanumerical) which are allotted to the parts to process the information (Ham et al., 1985). Parts are categorized based on significant attributes such as dimensions, type of material used, tolerance, operations required, basic shapes, surface finishing etc. In this approach, each part is assigned a code which is a string of numerical digits that stores information about the part. Singh and Rajamani (1996) demonstrated coding systems in their book and stated that the coding systems depict either hierarchical structure (monocode), or chain structure (polycode) or hybrid mode structure mixed with monocode and polycode.

Several part coding systems have been developed, e.g. Opitz (Opitz, 1970), MICLASS (TNO, 1975), DCLASS (Gallagher and Knight, 1985) and FORCOD (Jung and Ahluwalia, 1992) etc. Han and Ham (1986) have claimed that part families could be established more realistically by practicing the PCA due to the advantage of using the manufacturing and design attributes concurrently. Offodile reported a similarity metric based on the numeric codes for any pair of parts which could be utilized with an appropriate clustering method such as agglomerative clustering algorithm to form efficient part families (Offodile, 1992). Clustering analysis is practiced in Cellular Manufacturing System (CMS) as a competent methodology to facilitate the machine/part grouping problems.

Various machine/part grouping techniques are developed to solve manufacturing cell formation problems since last few decades, these include similarity coefficient methods, clustering analysis, array based techniques, graph partitioning methods etc. The similarity coefficient approach was first suggested by McAuley (1972). The basis of similarity coefficient methods is to calculate the similarity between each pair of machines and then to group the machines into cells based on their similarity measurements. Few studies have proposed to measure dissimilarity coefficients instead of similarity coefficient for machine-part grouping problems (Prabhakaran et al., 2002). Most of the similarity coefficient methods utilized machine–part mapping chart. Few of them are Single linkage clustering algorithm (McAuley, 1972), Average linkage clustering algorithm (Seifoddini and Wolfe, 1986).

Array based methods consider the rows and columns of the machine-part incidence matrix as the binary patterns and reconfigure them to obtain a block diagonal cluster formation. The rank order clustering algorithm is the most familiar array-based technique for cell formation (King, 1980). Substantial alterations and enhancements over rank order clustering algorithm have been described by King and Nakornchai (1982) and Chandrasekharan and Rajagopalan (1986). The direct clustering analysis technique has been implemented by Chan and Milner (1982).

Graph Theoretic Approach depicts that the machines as vertices and the similarity between machines as the weights on the arcs. Rajagopalan and Batra (1975) proposed the use of graph theory to form machine cells. Chandrasekharan and Rajagopalan (1986) proposed an ideal seed nonhierarchical clustering algorithm for cellular manufacturing. Graph searching algorithms was demonstrated by Ballakur and Steudel (1987) which select a crucial machine or part according to a pre-fixed criterion. Srinivasan (1994) implemented a method using minimum spanning tree (MST) for the machine-part cell formation problem.

This article proposes a novel part family identification technique namely SAPFOCS (Simulated Annealing for Part Family using Opitz Coding System) based on metaheuristic approach to investigate the nature of similarities among parts in the families and to describe the effectiveness of the techniques in solving the problem in hand.

## 2. Problem Definition

Opitz classification and coding system is used in this article which was proposed by Opitz (1970) at Aachen Technology University in West Germany. The basic code comprises of nine digits that can be extended by additional four digits. The general interpretations of the nine digits are as indicated in Fig. 1.The first 5 digits are called the form codes and describe the design or the general form of the part and hence aid in design retrieval process. Later, 4 more digits were added to the coding scheme in order to enhance the manufacturing information of the specific work part. These last four digits are called supplementary codes and respectively represent: dimensions, material, original shape of raw stock, and Accuracy of the work part.

The interpretation of first 9 digits are,
Digit 1: General shape of workpiece, otherwise called 'part-class'. This is further subdivided into rotational and non-rotational classes and further divided by size (length/diameter or length/width ratio.)
Digit 2: External shapes and relevant form. Features are recognized as stepped, conical, straight contours. Threads and grooves are also important.
Digit 3: Internal shapes. Features are solid, bored, straight or bored in stepped diameter. Threads and grooves are integral part.
Digit 4: Surface plane machining, such as internal or external curved surfaces, slots, splines.
Digit 5: Auxiliary holes and gear teeth.
Digit 6: Diameter or length of workpiece.
Digit 7: Material Used.
Digit 8: Shape of raw materials, such as round bar, sheet metal, casting, tubing etc.
Digit 9: Workpiece accuracy.

All the 9 digits are interpreted numerically (0-9). Examples of a mild steel forged round bar is shown in Fig. 2.

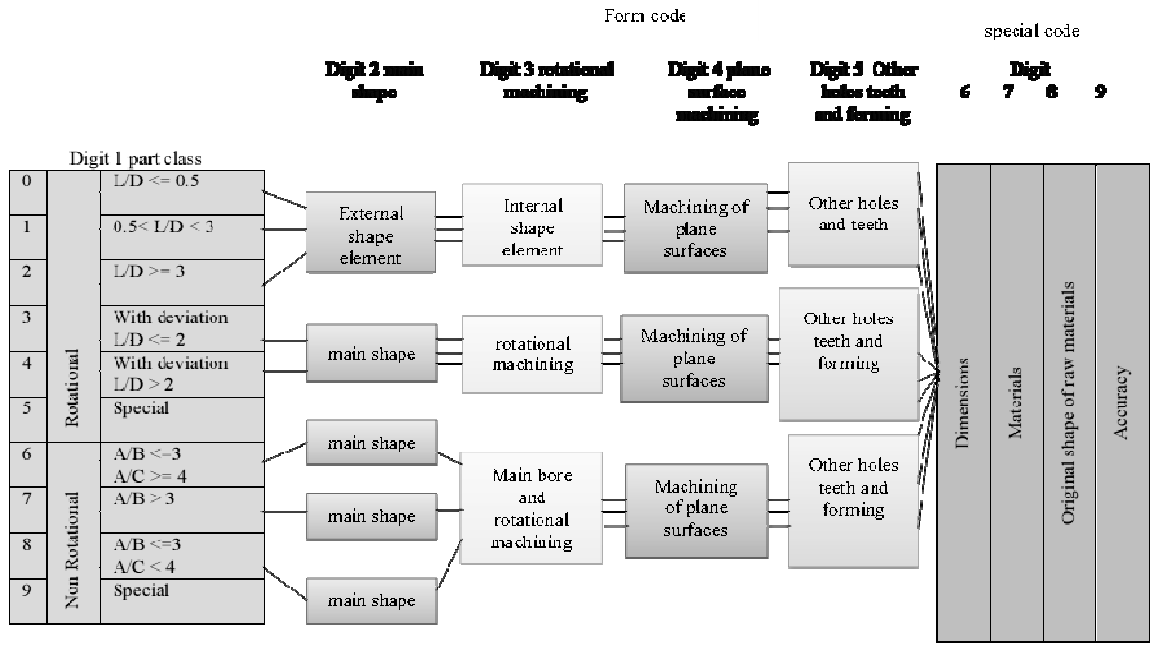

Fig. 1: Opitz part coding system

The Opitz codes of the round bar is 11103 2302 (Ham et al., 1985). The attributes are denoted as a1-a9 for the round bar,

a1=1 (Rotational parts, 0.5<L/D<3.)
a2=1 (External shape element, stepped to one end.)
a3=1 (Internal Shape element, smooth or stepped to one end.)
a4=0 (No surface machining.)
a5=5 (Auxiliary holes, radial.)
a6=2 (50 mm. < diameter <=100 mm.)
a7=3 (material is mild steel.)
a8=0 (Internal form: Round bar.)
a9=2 (Accuracy in coding digit.)

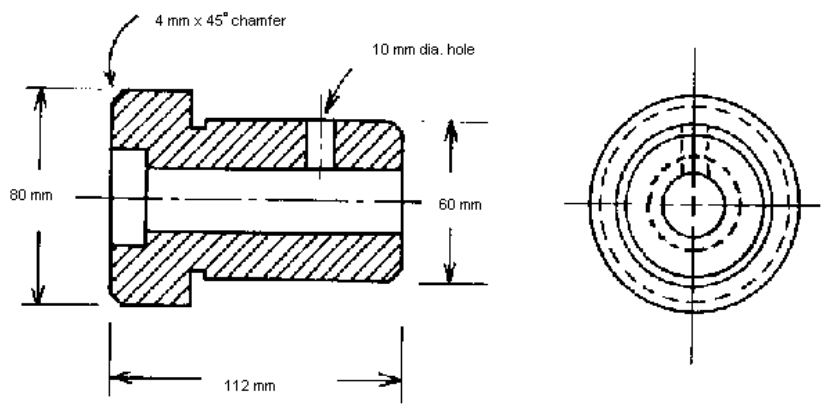

Fig. 2: Mild Steel Forged Round Rod

The part family formation problem stated in this research could be formulated using a part-attribute mapping matrix $B=[b_{ij}]$, of size $m \times n$, where $m$ is the number of parts and $n$ is the number of attributes of that part. $b_{ij}$ represents the coding value (0-9) of $j^{th}$ attribute of $i^{th}$ part. A 10×9 example problem based on Opitz coding system is shown in Fig. 3.

|     | a1 | a2 | a3 | a4 | a5 | a6 | a7 | a8 | a9 |
|-----|----|----|----|----|----|----|----|----|----|
| p1  | 0  | 0  | 1  | 0  | 0  | 9  | 1  | 3  | 6  |
| p2  | 0  | 0  | 1  | 0  | 1  | 2  | 6  | 5  | 6  |
| p3  | 0  | 0  | 2  | 0  | 0  | 6  | 2  | 1  | 7  |
| p4  | 0  | 0  | 2  | 3  | 0  | 4  | 1  | 6  | 9  |
| p5  | 0  | 0  | 3  | 0  | 0  | 8  | 0  | 3  | 5  |
| p6  | 0  | 1  | 1  | 0  | 0  | 7  | 7  | 8  | 9  |
| p7  | 0  | 1  | 1  | 0  | 2  | 4  | 4  | 8  | 1  |
| p8  | 0  | 1  | 2  | 3  | 0  | 0  | 9  | 5  | 9  |
| p9  | 0  | 2  | 1  | 0  | 4  | 8  | 4  | 4  | 2  |
| p10 | 0  | 2  | 1  | 3  | 0  | 4  | 4  | 8  | 2  |

Fig. 3: Problem #2 10×9 dataset

First 5 columns are form codes and rest of the 4 columns represent supplementary codes. The solution to the problem is to form the families of parts in such a way that the sum of similarities of parts in obtained part families would be maximized. Therefore metaheuristic algorithm is used in this article which groups the parts into families and maximizes the sum of similarities value.

## 3. Solution Methodology

During past few decades soft computing techniques are exhaustively practiced by researchers in the vicinity of CMS. Lee-Post (2000) proposed that GT coding system (DCLASS) could be efficiently used with simple Genetic Algorithm (GA) to cluster part families which is well suited for part design and process planning in production. Tavakkoli-Moghaddam et al. (2005) explained that dynamic condition of CFP becomes more complex and proposed Tabu Search (TS), Simulated Annealing (SA) and GA methods to solve this type of problems. Their study indicated that SA is better in terms of solution and complexity than TS, GA. Authors further (Tavakkoli-Moghaddam et al., 2008) introduced an integer programming model for dynamic CFP and implemented SA algorithm to obtain the optimal solutions. Das et al. (2006) proposed the multi-objective mixed integer-programming model for CMS design by minimizing machine operating and utilization cost and total material handling cost and maximizing system reliability. Lei and Wu (2006) worked with multi-objective cell formation (CF) problem and proposed a Pareto-optimality based on multi-objective tabu search (MOTS) with different objectives: minimization of the weighted sum of intercell and intracell moves and minimization of the total cell load variation. A hybrid methodology based on Boltzmann function from simulated annealing and mutation operator from GA was proposed by Wu et al. (2009) to optimize the initial cluster obtained from similarity coefficient method (SCM) and rank order clustering (ROC). Arkat et al. (2007) developed a sequential model based on SA for large-scale problems and compared their method with GA. Ateme-Nguema and Dao (2007) investigated an Ant Algorithm based TS heuristic for cellular system design problem (CSDP) and the methodology proved to be much quicker than traditional methods when considering operational sequence, time and cost. These Authors (2009) further proposed quantized Hopfield network for CFP to find optimal or near-optimal solution and TS was employed to improve the performance and the quality of solution of the network. Durán et al. (2010) reported a modified Particle Swarm algorithm with proportional likelihood instead of using velocity vector on CF problems where the objectives are the minimization of cell load variation and inter cellular parts movement and reported the stability of the method with low variability. A similar study was also performed by Anvari et al. (2010)

where a hybrid particle swarm optimization technique for CFP was reported. The initial solutions generated either randomly or using a diversification generation method and the technique also utilized mutation operator embedded in velocity update equation to avoid reaching local optimal solutions. Thereafter with due consideration, a wide variety of machine/part matrices were effectively solved by this approach.

Elaborated survey on metaheuristic based approaches in CMS could be obtained from Papaioannou and Wilson, (2010) and Ghosh et al., (2011). These studies demonstrated that PCA has merely been adopted to form part families using the stated methodologies. Therefore this article would explore an unexplored area of CMS by employing metaheuristic technique and part coding method.

### 3.1. Simulated Annealing (SA) Techniques

The SA algorithm simulates the physical annealing process, where particles of a solid arrange themselves into a thermal equilibrium. An introduction to SA can be found in the book by Aarts and Korst (1990). The standard type of applications concerns combinatorial optimization problems of the following form where $S$ is a finite set of feasible solutions.

$$\min_{x \in S} g(x) \qquad (1)$$

The algorithm uses a pre-defined neighbourhood structure on 'S'. A control parameter called *temperature* in analogy to the physical annealing process governs the search behaviour. In each iteration, a neighbour solution $y$ to the current solution $x$ is computed. If $y$ has a better objective function value than $x$, the solution $y$ is *accepted*, that is, the current solution $x$ is replaced by $y$. If, on the other hand, $y$ does not have a better objective function value than $x$, the solution $y$ is only accepted with a certain probability depending on (i) the difference of the objective function values in $x$ and $y$, and (ii) the temperature parameter. The pseudocode below demonstrates SA procedure.

Pseudocode (SA)
*initialize;*
*repeat*
*generate a candidate solution;*
*evaluate the candidate;*
*determine the current solution;*
*reduce the temperature;*
**until** *termination condition is met;*

### 3.2. Initial Solution Generation

The problem is to form the families of parts in such a way that the similarity among the parts in a same family would be maximized. In order to adopt any metaheuristic approach as a solution methodology, an initial solution should be generated quickly using some method. There could be many quick solution generation techniques available in literature such as similarity coefficient method (McAulay, 1972), Rank Order Clustering method (King, 1980) etc. Although every method has certain time complexities to be executed using some computer program, and moreover the generated initial solution is not essentially be the near optimal solution. Therefore in order to minimize the computational effort the initial solution is generated in this research using some random number generator. Thereafter metaheuristic approaches are applied in order to improve the quality of that generated solution gradually.

### 3.3. Initial Solution Representation

In this article an initial solution is represented using a bit string of length P, where P stands for the number of parts to be clustered. The initial solution for the problem of Figure 3 could be represented as '12121', which means assigning parts 1 to 5 to the cell 1,2,1,2 and 1 respectively. This initial solution is generated randomly for all three approaches, which might not be the best or near best solution to the problem. Therefore to understand the goodness of solution a fitness function is required.

### 3.4. Fitness Evaluation Function

The fitness value of each string is a measure of how well the part families are formed. The objective of part family formation is to maximize the sum of similarities of parts. Therefore, maximization of the sum of similarities could be utilized as the evaluation criteria to calculate the fitness of each solution string. This evaluation criteria is expressed mathematically as (Lee-Post, 2000),

$$Max\ f = \sum_{n=1}^{N} S_n \tag{1}$$

Where

$$S_n = \frac{\sum_{i \in n, j \in n} S_{ij}}{1 + C_2^{P_n}} \tag{2}$$

where $S_{ij}$ = similarity measure between part $i$ and part $j$
$C_2^{P_n}$ = Number of pairwise combinations formed in part family $n$, and $P_n$ is the number of parts in family $n$
definition of $S_{ij}$ is adopted from Offodile (1992) to accommodate numeric part coding, and is defined as follows:

$$S_{ij} = \frac{\sum_{k=1}^{K} S_{ijk}}{K} \tag{3}$$

Where

$$S_{ijk} = 1 - \frac{|b_{ik} - b_{jk}|}{R_k} \tag{4}$$

Where
$S_{ijk}$ is Similarity measured between part $i$ and part $j$ on attribute $k$,
$K$ is total number of attributes considered,
$b_{ik}$ is part coding for part $i$ on attribute $k$,
$b_{jk}$ is part coding for part $j$ on attribute $k$,
$R_k$ is range of possible part codings for all parts on attribute $k$.

Most important steps in proposed meta-heuristic techniques is the evaluation of the obtained solutions. In this step, the goodness (or fitness) of the solution is calculated, and based on the result, the solution may be deleted, kept, or marked as good. The TS algorithm maintains a tabu list to track the already evaluated solutions and tracks the ever best solution found in the process of execution. SA procedure also keeps the record of best solution encountered accordingly with the temperature reducing function. When a new solution is obtained, the goodness (fitness) function is applied, and based on the result, algorithms decide to add the solution to the elite list, or omit the solution and generate another one. During the TS and SA iterations, the goodness of each solution is calculated using equation (1) which is the objective function and performance measure as well.

Similarity coefficient based techniques are massively practiced in formation of manufacturing cells over the last few decades (Yin and Yasuda, 2005). In this article Complete Linkage (C-Linkage) technique is used along with a similarity metric method to generate a quick solution. When this technique develops the feasible solution to form initial part families, generation of initial solution is therefore accomplished.

*3.4.1. Similarity metric for part family formation*

In this research a similarity measure method based on part family identification technique is utilized (Offodile, 1992). It is presented as,

$$S_{ij} = \frac{\sum_{k=1}^{K} S_{ijk}}{K} \qquad (2)$$

Where

$$S_{ijk} = 1 - \frac{|b_{ik} - b_{jk}|}{R_k} \qquad (3)$$

Where

$S_{ijk}$ is Similarity measured between part $i$ and part $j$ on attribute $k$,
$K$ is total number of attributes considered,
$b_{ik}$ is part coding for part $i$ on attribute $k$,
$b_{jk}$ is part coding for part $j$ on attribute $k$,
$R_k$ is range of possible part codings for all parts on attribute $k$.

Therefore sum of similarities is derived using the formula,

$$f = \sum_{n=1}^{N} S_n \qquad (4)$$

Where

$$S_n = \frac{\sum_{i \in n, j \in n} S_{ij}}{0.001 + C_2^{P_n}} \qquad (5)$$

where $S_{ij}$ = similarity measure between part $i$ and part $j$
$C_2^{P_n}$ = Number of pair-wise combinations formed in part family $n$, and $P_n$ is the number of parts in family $n$ (in the denominator a small value of 0.001 is added to avoid the division by zero rule).

In order to facilitate the computation in Matlab the similarity matrix obtained using equation (2) further transformed into distance matrix using equation (5).

$$d_{ij} = 1 - S_{ij} \tag{6}$$

Complete Linkage Clustering Algorithm (C-Linkage) is adopted in present study to generate a feasible initial solution for the simulated annealing technique.

Abovementioned similarity metric technique is utilized to calculate the similarity coefficient value between pair of parts presented as rows of the part-attribute mapping matrix as given in Fig. 3. Thereafter the distance matrix is generated for (10×9) part-attribute incidence matrix using equation (6) is presented in Fig. 4.

*3.4.2. C-Linkage Clustering Techniques*

Linkage clustering method is mathematically simple algorithm practiced in hierarchical clustering analysis of data (McAuley, 1972). It delivers informative descriptions and visualization of possible data clustering structures. When there exists hierarchical relationship in data this approach can be more competent. C-Linkage is also identified as furthest neighbor method which uses the maximum distance between two clusters. The distance between cluster *i* and another cluster *j* is defined as:

$$C_{ij} = \max(d_{ij}), i \in (1, \ldots, n_r), j \in (1, \ldots, n_s) \tag{7}$$

Using equation (7), an intermediate matrix could be obtained which is a *(p-1)×3* matrix, where *p* is the number of parts in the original dataset. Columns of the matrix contain cluster indices linked in pairs to form a binary tree. The leaf nodes are numbered from *1 to p*. Leaf nodes are the singleton clusters from which all higher clusters are built. Further the dendrogram could be obtained from the matrix which indicates a tree of potential solution. An example dendrogram structure obtained for the parts using C-Linkage method is shown in Fig. 5, which depicts the clear cluster information of part families.

|     | p1 | p2 | p3 | p4 | p5 | p6 | p7 | p8 | p9 | p10 |
|-----|----|----|----|----|----|----|----|----|----|----|
| p1  | 0 | 0.185185 | 0.098765 | 0.185185 | 0.061728 | 0.209877 | 0.259259 | 0.333333 | 0.185185 | 0.271605 |
| p2  |   | 0 | 0.185185 | 0.197531 | 0.222222 | 0.17284 | 0.17284 | 0.17284 | 0.222222 | 0.209877 |
| p3  |   |   | 0 | 0.160494 | 0.111111 | 0.209877 | 0.259259 | 0.283951 | 0.234568 | 0.271605 |
| p4  |   |   |   | 0 | 0.197531 | 0.197531 | 0.246914 | 0.17284 | 0.320988 | 0.185185 |
| p5  |   |   |   |   | 0 | 0.246914 | 0.271605 | 0.345679 | 0.197531 | 0.283951 |
| p6  |   |   |   |   |   | 0 | 0.197531 | 0.197531 | 0.246914 | 0.209877 |
| p7  |   |   |   |   |   |   | 0 | 0.320988 | 0.148148 | 0.08642 |
| p8  |   |   |   |   |   |   |   | 0 | 0.37037 | 0.259259 |
| p9  |   |   |   |   |   |   |   |   | 0 | 0.185185 |
| p10 |   |   |   |   |   |   |   |   |   | 0 |

Fig. 4: distance matrix obtained from (10×9) part-attribute incidence matrix of Fig. 3

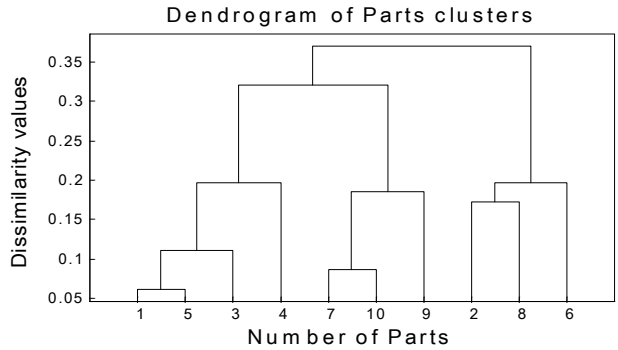

Fig. 5: Dendrogram of part clusters of 10×9 example problem #2

From Figure 5 it can be stated that 3 part families are obtained using the hierarchical notations. The part families are, Family 1 {7,9,10}, Family 2 {1,3,4,5}, Family 3 {2,6,8}.

The undermentioned pseudocode would explain the hierarchical C-Linkage method,

---

Input: part-attribute incidence matrix A
*Step 1. Procedure similarity()*
*Step 1.1. Compute similarity values between pair of machines using equation (2)*
*Step 1.2. Compute the similarity matrix of the parts*
*Step 1.3. transform the similarity matrix into a distance matrix using equation (6)*
*Step 1.4. End*
*Step 2. Procedure Cluster()*
*Step 2.1. loop*
*Step 2.2. Compute the furthest Euclidian distance between two clusters for C-Linkage method*
*Step 2.3. Construct matrices of size (m-1)×3 to from the hierarchical tree structure*
*Step 2.4. Construct the dendrogram*
*Step 2.5. loop*
*Step 2.6. create part families for the highest level of similarity coefficient*
*Step 2.7. End*
Output: initial part family configurations

---

### 3.5. The Proposed SAPFOCS Technique

This subsection describes the proposed algorithm SAPFOCS in depth. In this algorithm, the number of part families resulting in the best solution is fixed initially. However, the flexibility is conserved for users to specify the number of cells they prefer to maximize the percentage of perfections of the obtained part groups. The solution obtained from C-Linkage method is given as an input to the SAPFOCS which is presented as a bit string (instead of the solution matrix) of length $p$ ($p$ is the number of parts) in order to minimize the computational effort. The initial input string for the example problem of Fig. 3 could be symbolized as '2322231311', which means assigning parts 7, 9, 10 to the family 1, parts 1, 3, 4, 5 to family 2 and parts 2, 6, 8 to family 3 respectively as stated by C-Linkage method. To understand the goodness of the solution a performance evaluation criteria is assumed to be explained which is presented by equation (4). Therefore the objective is set to maximize the $f$ value of the objective function. Some symbolizations used in the SAPFOCS algorithm are introduced as,

$S_{cur}$ → current solution
$S_i$ → neighbourhood solution
$S_{best}$ → best solution found so far
$T_{init}$ → initial temperature
$T_{final}$ → freezing temperature
$α$ → temperature reducing factor
$M$ → Markov chain length
$iter$ → iteration number
$f_i$ → current objective value
$f_{best}$ → best objective value

The steps of the proposed algorithm SAPFOCS can be summarized as follows.

*Algorithm SAPFOCS*
*Step 1. Obtain an initial solution $S_0$ by using similarity coefficient and C-Linkage procedures in Section 3.2.*
*Step 2. Evaluate $S_0$ and Calculate corresponding objective value $f_0$. $f_0 = f(S_0)$*
*Step 3.* **Set** *$f_{best} = f_0$,* **Set** *$S_{best} = S_0 = S_{cur}$.*
*Step 4. Initialize SAPFOCS and its parameters: $T_{init}$, $T_{final}$, α, M, iter = 0, count=0, count1=0.*
*Step 5.* **If** *count < M, then repeat Steps 5.1 to 5.9.*
*Step 5.1. Generate a new part family formation configuration neighbourhood searching by performing single-move (randomly selecting a part and moving it to another family).*
*Step 5.2. Read part family formation configuration from above steps and generate corresponding neighbourhood solution $S_i$.*
*Step 5.3.* **If** *$f(S_i) > f_{best}$, then $S_{best} = S_i$, $S_{cur} = S_i$, count = count + 1,* **go to** *Step 5.*
*Step 5.4.* **If** *$f(S_i) = f_{best}$, then $S = S_i$, count1= count1 + 1, count = count + 1,* **go to** *Step 5.*
*Step 5.5. Compute $δ = f(S_i) - f(S_{cur})$. Obtain a random variable r in the range of U(0,1).*
*Step 5.6.* **If** *$e^{δ/T} > r$,*
*Step 5.7. set $S_{cur} = S_i$,*
*Step 5.8. count1= 0;*
*Step 5.9.* **else** *count1 = count1 + 1.*
*Step 5.10. iter = iter + 1.*
*Step 5.11.* **until** *freezing temperature ($T_{final}$) is reached;*
*Step 5.12. reduce the temperature using $T_i = α × T_{i-1}$ function;*

The SAPFOCS consists of an SA procedure that is repetitively employed until a part family configuration is achieved which attains the highest sum of similarities amongst the part families. Initially number of part families is set to a prefixed number (the convention is used as 4 parts to be placed in one family). All the parameters and counters are initialized in step 4. A special move, namely single-move, is utilized in the proposed algorithm to guide the solution searching procedure. From the understanding of exhaustive testing, it is spotted that single move ordinarily leads to improved solutions effortlessly and competently. Thus single-move is practiced as a principle component for finding better neighbourhood solution in step 5.1. SAPFOCS also verifies the number of instances when neighbourhood solutions become static. If this number attains a pre-fixed constant value, the objective value of current configuration is compared to the optimal solution obtained thus far to conclude whether to prolong the iterations or stop with the best solution achieved.

## 4. Experiments and Discussion

In order to employ the proposed SAPFOCS as a solution methodology to solve the part family formation problem, the effects of changing the values of the various parameters are studied. Determining the optimal set of parameters are crucial in this regards. Therefore in this article the Taguchi's orthogonal design method is employed to determine the optimal values of the parameters.

### 4.1. Taguchi Method for Parameters Selection

The parameters are Initial temperature ($T_{init}$), temperature reducing factor ($\alpha$) and Markov chain length ($M$) (Other parameter such as final temperature ($T_{final}$) is taken as constant value = 0.002 initially). The parameters are termed as factors, and each factor has three discrete levels (Table 1). Hence an L9 orthogonal array is used, and this recommends that 9 sets of Taguchi experiments are prerequisite and the results are evaluated by using an analysis of variance (ANOVA) technique. The parameter settings for each experiment are shown in Table 2.

Table 1. Levels of parameters tested

| levels | Parameters | | |
|---|---|---|---|
| | $M$ | $\alpha$ | $T_{init}$ |
| 1 | 20 | 0.75 | 10 |
| 2 | 30 | 0.85 | 20 |
| 3 | 40 | 0.95 | 30 |

Table 3 presents the results of the corresponding ANOVA analysis with S/N ratio (Larger-the-better). In Table 3, the variance ratios ($F$ ratios) of the factors are determined. A test of significance at 95% confidence level is employed to spot the significance of these factors. The $F$ values of the factors $T_{init}$, $\alpha$, $M$ are investigated to be less than the critical level with degrees of freedom at (2, 8). This suggests that all the selected parameters are significant factors in the proposed approach. The response table (Table 4) depicts the average of each response characteristic for each level of each of the factors. The tables include ranks based on Delta ($\delta$) statistics, which compare the relative magnitude of effects. The Delta statistic states the difference between the largest and the smallest average for each factor. Ranks are assigned based on Delta values. Using the level averages in the response table optimal set of levels of the factors could be determined which yields the best result.

In present study, the ranks indicate that Markov chain length ($M$) has the greatest influence. Initial temperature ($T_{init}$) has the next greatest influence, followed by temperature reducing factor ($\alpha$).

The objective of the simulated annealing is to maximize the value of objective function of equation (4), therefore factor levels should be fixed in such a way that the highest objective value could be achieved. The level averages in the response table show that the optimal solution is obtained when $T_{init}$ is set to 30, $\alpha$ is set to 0.85 and $M$ is set to at 40. Which could further be confirmed by the main effects plot of Fig. 6.

Table 2. The Experimental Settings of the Taguchi Experiments

| Experiments | $T_{init}$ | α | M | responses |
|---|---|---|---|---|
| 1 | 10 | 0.75 | 20 | 5.44289 |
| 2 | 10 | 0.85 | 30 | 5.4903 |
| 3 | 10 | 0.95 | 40 | 5.50745 |
| 4 | 20 | 0.75 | 30 | 5.45195 |
| 5 | 20 | 0.85 | 40 | 5.49802 |
| 6 | 20 | 0.95 | 20 | 5.524 |
| 7 | 30 | 0.75 | 40 | 5.54252 |
| 8 | 30 | 0.85 | 20 | 5.52575 |
| 9 | 30 | 0.95 | 30 | 5.46335 |

Table 3. ANOVA table

| Factors | Degrees of Freedom | Factor Sum of squares | Mean Square (Variance) | F Ratio (Variance) | % contribution |
|---|---|---|---|---|---|
| $T_{init}$ | 2 | 0.001412 | 0.000706 | 0.36 | 0.733 |
| α | 2 | 0.001062 | 0.000531 | 0.27 | 0.785 |
| M | 2 | 0.003435 | 0.001717 | 0.88 | 0.531 |
| Residual Error | 2 | 0.003884 | 0.001942 | | |
| Total | 8 | 0.009793 | | | |

Table 4. Response table

| Levels | $T_{init}$ | α | M |
|---|---|---|---|
| 1 | 5.480 | 5.479 | 5.498 |
| 2 | 5.491 | 5.505 | 5.469 |
| 3 | 5.511 | 5.498 | 5.516 |
| δ | 0.030 | 0.026 | 0.047 |
| Rank | 2 | 3 | 1 |

## 4.2. Computational Results

The proposed SAPFOCS technique is coded in Matlab 7.0 platform on a PIV core 2 duo laptop computer and tested on 5 different problem datasets of size 5×9 to 27×9. The largest dataset has been obtained from Haworth (1968) using Opitz coding system. Remaining 4 problems are designed using the aforestated coding system. Problem datasets are provided in Fig. 3 and Fig. 7 to 10. The obtained results are compared and shown in Table 5.

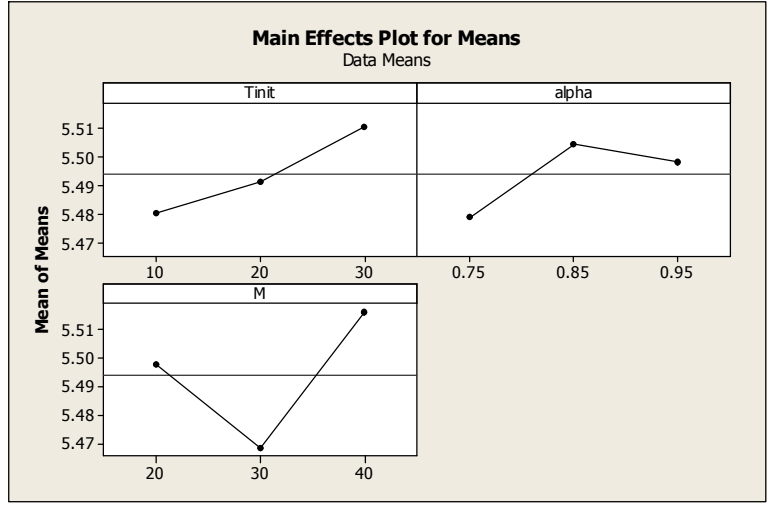

Fig. 6: Main effects plot

```
     a1  a2  a3  a4  a5  a6  a7  a8  a9
p1   0   0   1   0   0   9   1   3   6
p2   0   0   1   0   1   2   6   5   6
p3   0   0   2   0   0   6   2   1   7
p4   0   0   2   3   0   4   1   6   9
p5   0   0   3   0   0   8   0   3   5
```

Fig. 7: Problem #1 (5×9) dataset

```
     a1  a2  a3  a4  a5  a6  a7  a8  a9
p1   0   0   1   0   0   9   1   3   6
p2   0   0   1   0   1   2   6   5   6
p3   0   0   2   0   0   6   2   1   7
p4   0   0   2   3   0   4   1   6   9
p5   0   0   3   0   0   8   0   3   5
p6   0   1   1   0   0   7   7   8   9
p7   0   1   1   0   2   4   4   8   1
p8   0   1   2   3   0   0   9   5   9
p9   0   2   1   0   4   8   4   4   2
p10  0   2   1   3   0   4   4   8   2
p11  0   2   2   0   4   6   8   8   8
p12  0   2   3   0   4   7   5   6   7
p13  1   0   0   0   0   9   2   8   1
p14  1   0   1   0   0   7   6   6   0
p15  1   1   0   0   0   1   8   3   8
```

Fig. 8: Problem #3 (15×9) dataset

Convergence analysis are almost equivalent for all the problem datasets. Problem #5 of size 27×9 is selected as an example to demonstrate the convergence curve during iterations of the proposed metaheuristic technique as presented in Fig. 11. For the first iteration the objective function $f$ attained a value of 5.244. Since the computer program is designed to maximize the objective function with the iteration counts therefore at $3^{rd}$ iteration it attained the value of 5.431, an increase of 3.56%. The final optimal solution is obtained during the $11^{th}$ iteration having the objective value of 5.537, an increase of 5.58%. Based on the experimentation for all the datasets reported in this article, it is observed that the objective value is increased with the iteration counts till it reaches the best objective value at some iteration and thereafter the objective value continues to remain constant even if the number of iterations is increased. Since the proposed metaheuristic algorithm gives the same pattern of convergence for all the tested problems therefore the convergence property is established. For the 27×9 problem the SAPFOCS approach is executed for 38 iterations. This proposed algorithm took 11.1908 CPU seconds to attain the best solution which proves its computational efficiency. Fig. 11 depicts the maximum fitness value achieved for the solutions in each iteration along with the mean fitness value achieved.

|     | a1 | a2 | a3 | a4 | a5 | a6 | a7 | a8 | a9 |
|-----|----|----|----|----|----|----|----|----|----|
| p1  | 0  | 0  | 1  | 0  | 0  | 9  | 1  | 3  | 6  |
| p2  | 0  | 0  | 1  | 0  | 1  | 2  | 6  | 5  | 6  |
| p3  | 0  | 0  | 2  | 0  | 0  | 6  | 2  | 1  | 7  |
| p4  | 0  | 0  | 2  | 3  | 0  | 4  | 1  | 6  | 9  |
| p5  | 0  | 0  | 3  | 0  | 0  | 8  | 0  | 3  | 5  |
| p6  | 0  | 1  | 1  | 0  | 0  | 7  | 7  | 8  | 9  |
| p7  | 0  | 1  | 1  | 0  | 2  | 4  | 4  | 8  | 1  |
| p8  | 0  | 1  | 2  | 3  | 0  | 0  | 9  | 5  | 9  |
| p9  | 0  | 2  | 1  | 0  | 4  | 8  | 4  | 4  | 2  |
| p10 | 0  | 2  | 1  | 3  | 0  | 4  | 4  | 8  | 2  |
| p11 | 0  | 2  | 2  | 0  | 4  | 6  | 8  | 8  | 8  |
| p12 | 0  | 2  | 3  | 0  | 4  | 7  | 5  | 6  | 7  |
| p13 | 1  | 0  | 0  | 0  | 0  | 9  | 2  | 8  | 1  |
| p14 | 1  | 0  | 1  | 0  | 0  | 7  | 6  | 6  | 0  |
| p15 | 1  | 1  | 0  | 0  | 0  | 1  | 8  | 3  | 8  |
| p16 | 1  | 1  | 1  | 0  | 0  | 4  | 0  | 2  | 1  |
| p17 | 1  | 2  | 0  | 3  | 0  | 9  | 6  | 3  | 2  |
| p18 | 2  | 0  | 0  | 0  | 0  | 9  | 3  | 5  | 6  |
| p19 | 2  | 3  | 0  | 0  | 0  | 4  | 8  | 7  | 2  |
| p20 | 2  | 5  | 0  | 0  | 0  | 8  | 5  | 3  | 4  |

Fig. 9: Problem #4 (20×9) dataset

|     | a1 | a2 | a3 | a4 | a5 | a6 | a7 | a8 | a9 |
|-----|----|----|----|----|----|----|----|----|----|
| p1  | 0  | 0  | 1  | 0  | 0  | 9  | 1  | 3  | 6  |
| p2  | 0  | 0  | 1  | 0  | 1  | 2  | 6  | 5  | 6  |
| p3  | 0  | 0  | 2  | 0  | 0  | 6  | 2  | 1  | 7  |
| p4  | 0  | 0  | 2  | 3  | 0  | 4  | 1  | 6  | 9  |
| p5  | 0  | 0  | 3  | 0  | 0  | 8  | 0  | 3  | 5  |
| p6  | 0  | 1  | 1  | 0  | 0  | 7  | 7  | 8  | 9  |
| p7  | 0  | 1  | 1  | 0  | 2  | 4  | 4  | 8  | 1  |
| p8  | 0  | 1  | 2  | 3  | 0  | 0  | 9  | 5  | 9  |
| p9  | 0  | 2  | 1  | 0  | 4  | 8  | 4  | 4  | 2  |
| p10 | 0  | 2  | 1  | 3  | 0  | 4  | 4  | 8  | 2  |
| p11 | 0  | 2  | 2  | 0  | 4  | 6  | 8  | 8  | 8  |
| p12 | 0  | 2  | 3  | 0  | 4  | 7  | 5  | 6  | 7  |
| p13 | 1  | 0  | 0  | 0  | 0  | 9  | 2  | 8  | 1  |
| p14 | 1  | 0  | 1  | 0  | 0  | 7  | 6  | 6  | 0  |
| p15 | 1  | 1  | 0  | 0  | 0  | 1  | 8  | 3  | 8  |
| p16 | 1  | 1  | 1  | 0  | 0  | 4  | 0  | 2  | 1  |
| p17 | 1  | 2  | 0  | 3  | 0  | 9  | 6  | 3  | 2  |
| p18 | 2  | 0  | 0  | 0  | 0  | 9  | 3  | 5  | 6  |
| p19 | 2  | 3  | 0  | 0  | 0  | 4  | 8  | 7  | 2  |
| p20 | 2  | 5  | 0  | 0  | 0  | 8  | 5  | 3  | 4  |
| p21 | 7  | 0  | 0  | 0  | 3  | 0  | 7  | 8  | 0  |
| p22 | 7  | 0  | 0  | 3  | 3  | 3  | 4  | 5  | 9  |
| p23 | 7  | 0  | 0  | 5  | 3  | 8  | 3  | 3  | 5  |
| p24 | 7  | 0  | 0  | 6  | 3  | 0  | 1  | 7  | 4  |
| p25 | 2  | 0  | 8  | 0  | 1  | 1  | 1  | 5  | 5  |
| p26 | 1  | 5  | 1  | 0  | 0  | 2  | 6  | 4  | 3  |
| p27 | 6  | 5  | 4  | 4  | 3  | 6  | 0  | 7  | 0  |

Fig. 10: Problem #5 (27×9) dataset

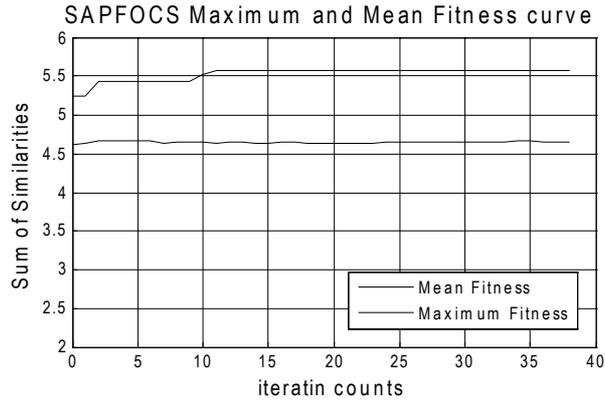

Fig. 11: Convergence curve of SAPFOCS

The proposed technique is thus compared with C-Linkage algorithm and results are presented in Table 5 which demonstrates that the proposed methodology is competent to attain better solution than C-Linkage method and the optimal parameters setting enhances the effectiveness of the technique. Despite of the fact that the solutions obtained are not identical in all instances therefore the sum of similarities are not identical for the test datasets. It further depicts that SAPFOCS technique outperformed the C-Linkage method and obtained 80% improved solutions.

In terms of computational time the proposed SAPFOCS method is efficient and took minimum CPU time (not more than 12 CPU seconds) for all the datasets tested as reported in Table 5. Table 6 reports the percentage of perfection achieved while developing the part groups by SAPFOCS and compares successfully with C-Linkage solutions. Further Fig. 12 establishes the superiority shown by SAPFOCS over C-Linkage technique. For the largest problem tested SAPFOCS achieved nearly 33.9% improved solution than the C-Linkage technique.

## 5. Conclusions

A state-of-the-art metaheuristic algorithm, SAPFOCS is proposed and employed in this research to form part families. Since part coding and classification techniques are merely adopted in group technology problems, therefore the objective of this study is to utilize the part coding system based on Opitz part coding. 5 test datasets ranging from 5×9 to 27×9 are tested using the aforementioned technique. Due to the NP-complete nature of the reported problems this method is highly effective to produce optimal solutions. Parameters selection is a critical decision for the simulated annealing to be able to obtain near optimal solutions, therefore Taguchi's orthogonal design method is employed to obtain optimal set of parameters for SAPFOCS algorithm. The proposed method is then compared with C-Linkage technique successfully. The objective function utilized in this study is to maximize the sum of similarities among parts in all the part families formed. From the foundation of this research work the number of part families to be formed is considered as constant. As shown in Table 6 SAPFOCS outperformed C-Linkage algorithm in terms of solution quality (sum of similarities value) which is further depicted in Fig. 12 in terms of the percentage of perfections achieved while forming the part families by each of the methods. This study has assumed identical weightage for each and every attribute, however in formation of part families some attributes could be more significant than the other attributes. Therefore future work could be done by considering fractional weightage for each of the attributes. This work could also be extended by considering operational time and sequence of each part in more complex and realistic environment to develop more effective and robust part families.

Table 5. Comparison of performance shown by SAPFOCS and C-Linkage techniques

| # | Dataset size | Part families obtained | | Maximum similarities | | CPU time for SAPFOCS (in Seconds.) |
|---|---|---|---|---|---|---|
| | | *C-Linkage* | *SAPFOCS* | *C-Linkage* | *SAPFOCS* | |
| 1 | 5×9 | Family 1 {2}, Family 2 {1,3.4,5} | Family 1 {2,3,4}, Family 2 {1,5} | 0.8641 | 1.75599 | 3.5674 |
| 2 | 10×9 | Family 1 {7,9,10}, Family 2 {1,3,4,5}, Family 3 {2,6,8} | Family 1 {2,3,4,6,8,9}, Family 2 {1, 5}, Family 3 {7,10} | 2.5425 | 2.6269 | 4.6915 |
| 3 | 15×9 | Family 1 {6, 11, 12}, Family 2 {2, 8, 15}, Family 3 {1, 3, 4, 5}, Family 4 {7, 9, 10, 12, 13, 14} | Family 1 {3,5}, Family 2 {7,9,12}, Family 3 {1, 2, 4, 8, 10, 13, 14, 15}, Family 4 {6, 11} | 3.4338 | 3.45274 | 6.2381 |
| 4 | 20×9 | Family 1 {7,10,14,19}, Family 2 {9, 13, 17, 18, 20}, Family 3 {6, 11, 12}, Family 4 {2, 8, 15}, Family 5 {1, 3, 4, 5, 16} | Family 1 {7,10,14,19}, Family 2 {9, 13, 17, 18, 20}, Family 3 {6, 11, 12}, Family 4 {2, 8, 15}, Family 5 {1, 3, 4, 5, 16} | 4.2510 | 4.2510 | 8.6916 |
| 5 | 27×9 | Family 1 {6,11,12}, Family 2 {2,8,15}, Family 3 {21}, Family 4 {22,23,24}, Family 5 {27}, Family 6 {1, 3, 4, 5, 16, 25}, Family 7 {7, 9, 10, 13, 14, 17, 18, 19, 20, 26} | Family 1 {1,5,16,25}, Family 2 {6,13}, Family 3 {7,10}, Family 4 {9,17}, Family 5 {8,11,15,21}, Family 6 {3,4,18,20,24, 27}, Family 7 {2,12,14,19,22, 26} | 4.1631 | 5.53717 | 11.1908 |

Table 6. Comparison among percentage of perfections of the obtained results by the SAPFOCS and C-Linkage techniques

| Problem Dataset | No. of part families formed(N) | Sum of similarities, $\sum S_n$ | | Perfection percentage, $\sum S_n / N \times 10$ | |
|---|---|---|---|---|---|
| | | *C-Linkage* | *SAPFOCS* | *C-Linkage* | *SAPFOCS* |
| 1 | 2 | 0.8641 | 1.75599 | 43.20 | 87.80 |
| 2 | 3 | 2.5425 | 2.6269 | 84.75 | 87.56 |
| 3 | 4 | 3.4338 | 3.45274 | 85.84 | 86.34 |
| 4 | 5 | 4.2510 | 4.2510 | 85.02 | 85.02 |
| 5 | 7 | 4.1631 | 5.53717 | 59.47 | 79.10 |

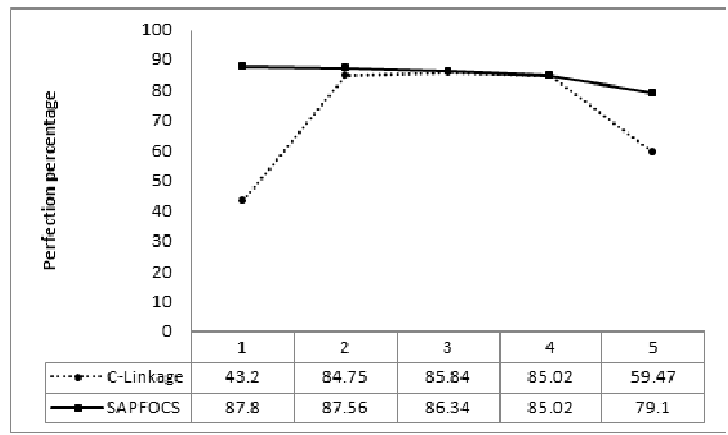

Fig. 12: Dominance shown by SAPFOCS over C-Linkage